\documentclass[letterpaper,10pt,journal,twoside]{IEEEtran}

\usepackage[T1]{fontenc}
\usepackage{textcomp}

\usepackage{amsmath}
\usepackage{amssymb}
\usepackage{bbm}
\usepackage{dsfont}
\usepackage{siunitx}
\usepackage{pifont}

\usepackage[pdftex]{graphicx}
\usepackage{svg}
\usepackage{tikz}
\usetikzlibrary{spy}
\usepackage{adjustbox}

\usepackage{array}
\usepackage{booktabs}
\usepackage{multirow}
\usepackage{makecell}
\usepackage{diagbox}

\usepackage{algorithm}
\usepackage{algpseudocode}

\usepackage{xspace}
\usepackage{balance}
\usepackage{romannum}
\usepackage{xparse}
\usepackage{etoolbox}
\usepackage{lipsum}

\usepackage{xcolor}
\usepackage{colortbl}
\usepackage{soul}

\usepackage[font=small]{caption}
\usepackage{xurl}
\usepackage[hidelinks]{hyperref}
\usepackage{cleveref}

\makeatletter
\def\bstctlcite#1{%
  \@bsphack
  \@for\@citeb:=#1\do{%
    \edef\@citeb{\expandafter\@firstofone\@citeb}%
    \if@filesw\immediate\write\@auxout{\string\citation{\@citeb}}\fi
  }%
  \@esphack
}
\makeatother

\crefname{figure}{Fig.}{Figs.}
\Crefname{figure}{Fig.}{Figs.}

\usepackage{rpm_acronyms}
\usepackage{rpm_math}
\usepackage{rpm_misc}

\newcommand{\cmtt}[1]{{\fontfamily{cmtt}\selectfont #1}}

\newcommand{\xmark}{\ding{55}}

\definecolor{lightgreen}{RGB}{200,255,200}
\definecolor{lightorange}{RGB}{252,210,153}
\definecolor{lightyellow}{RGB}{255,255,180}
\definecolor{lightpink}{RGB}{255,182,193}

\definecolor{darkblue}{RGB}{0,0,139}
\definecolor{darkgreen}{RGB}{10,106,71}

\definecolor{condzero}{RGB}{210,235,210}
\definecolor{condone}{RGB}{230,210,120}
\definecolor{condtwo}{RGB}{235,170,85}
\definecolor{condthree}{RGB}{225,120,65}
\definecolor{condfour}{RGB}{210,80,80}

\sethlcolor{red!50}

\begin{document}

\title{Hilti-Trimble-Oxford Dataset:\\360 Visual-Inertial Benchmark with Floor Plan Priors for SLAM and Localization}

\author{
Samuele Centanni$^{1*}$,
Yuhao Zhang$^{2*}$,
Yifu Tao$^{2}$,
Julien Kindle$^{1,3}$,
Frank Neuhaus$^{4}$,
Tilman Koß$^{4}$,\\
Aryaman Patel$^{5}$,
Michael Helmberger$^{1}$,
Emilia Szyma\'nska$^{1}$,
Torben Gr\"aber$^{1}$,
and Maurice Fallon$^{2}$%
\thanks{Manuscript received: July 7, 2026; Revised: September 22, 2026; Accepted: October 11, 2026.
This paper was recommended for publication by Editor Hyun Myung upon evaluation of the Associate Editor and Reviewers’ comments.
\textit{($^*$Samuele Centanni and Yuhao Zhang contributed equally to this work.)}
\textit{(Corresponding author: Emilia Szymańska.)}
}%
\thanks{$^{1}$Samuele Centanni, Julien Kindle, Michael Helmberger, Emilia Szyma\'nska, and Torben Gr\"aber are with Hilti AG, Corporate Research \& Technology, Schaan, Liechtenstein (e-mail: \cmtt{emilia.szymanska@hilti.com}).}%
\thanks{$^{2}$Yuhao Zhang, Yifu Tao, and Maurice Fallon are with Dynamic Robot Systems Group, University of Oxford, Oxford, U.K. (e-mail: \cmtt{yuhao@robots.ox.ac.uk}).}%
\thanks{$^{3}$Julien Kindle is also with Robotics Systems Lab, ETH Zürich, Zürich, Switzerland (e-mail: \cmtt{jkindle@ethz.ch}).}%
\thanks{$^{4}$Frank Neuhaus and Tilman Koß are with Vision \& Robotics GmbH, Koblenz, Germany (e-mail: \cmtt{frank.neuhaus@vision-robotics.de}).}%
\thanks{$^{5}$Aryaman Patel is with Trimble Inc., Denver, USA (e-mail: \cmtt{aryaman\_patel@trimble.com}).}%
\thanks{Dataset and benchmark: \cmtt{\textcolor{blue}{https://hilti-trimble-challenge.com}}}%
\thanks{Digital Object Identifier (DOI): see top of this page.}

}

\markboth{IEEE Robotics and Automation Letters. Preprint Version. Accepted October, 2026}%
{Centanni \MakeLowercase{\textit{et al.}}: Hilti-Trimble-Oxford Dataset}

\IEEEpubid{%
  \parbox[t]{\textwidth}{%
    \footnotesize\centering
    \copyright~2026 IEEE. Personal use of this material is permitted.  Permission from IEEE must be obtained for all other uses, in any current or future media, including reprinting/republishing this material for advertising or promotional purposes, creating new collective works, for resale or redistribution to servers or lists, or reuse of any copyrighted component of this work in other works.
  }%
}

\maketitle
\bstctlcite{IEEEexample:BSTcontrol}

\begin{abstract}
Automated progress monitoring on construction sites is an active area of research and development. Robot and human-carried mapping systems have been developed to build 3D maps of building and infrastructure projects. While LiDAR-based mapping systems achieve high accuracy, the cost of LiDAR can be prohibitive. Consumer-grade cameras with wide field of view (\textit{``360 cameras''}) combined with embedded inertial measurement units (IMUs) provide a cost-effective alternative. To support change detection and progress monitoring, highly accurate visual simultaneous localization and mapping (SLAM) and floor plan localization systems are required. In this paper, we present a high-quality dataset collected at an active construction site, which captures realistic challenges such as variable lighting conditions, moving workers, fast motions, and repetitive structures. The dataset offers thirty visual-inertial sequences recorded across ten floors over an eight-month period of the construction project. Ground truth trajectories are collected using a high quality LiDAR-inertial SLAM system rigidly attached to the 360 camera. Additionally, we report the results of an open research challenge evaluating the best visual SLAM and localization systems from around the world.
The Challenge attracted substantially higher participation in SLAM, with 62 teams compared to 22 in floor plan localization, which suggests that this task receives less attention from researchers and motivates further exploration through proposed dataset and challenge.
\end{abstract}


\section{Introduction}
\label{sec:intro}

\IEEEPARstart{T}{he} buildings and construction industry accounts for approximately $11$--$13\,\%$ of global GDP \cite{UNEP_GlobalABC_2026_BuildingsGSR}. It is, however, a sector that has seen minimal improvements to productivity, in part due to the limited digitization of work processes, rework due to defects and quality deviations, occupational safety hazards, and demographic changes in the construction workforce~\cite{armeni2024tc_article}. These challenges motivate automated methods for tracking the progress of construction to provide site managers with timely insights into the state of the project.

A common approach to construction site monitoring is the regular mapping of the environment using a variety of sensors  such as LiDAR, RGB cameras, or terrestrial laser scanners~\cite{bosche2009cadlaser,turkan2012progress}. These sensors may be operated manually or mounted on robotic platforms, including legged robots, drones, and wheeled platforms~\cite{halder2021quadruped,arjmand2023roboticlaser}. The usefulness of the resulting maps depends on the extent to which they can support change detection across consecutive mapping sessions or comparison against architectural models~\cite{bosche2015scantobim}. Detecting errors or deviations early can lead to substantial savings in both cost and time. The maps can also serve as documentary evidence of the building project~\cite{BSR_GoldenThread}.

\begin{figure}[t!]
    \centering
    \includegraphics[width=\linewidth]{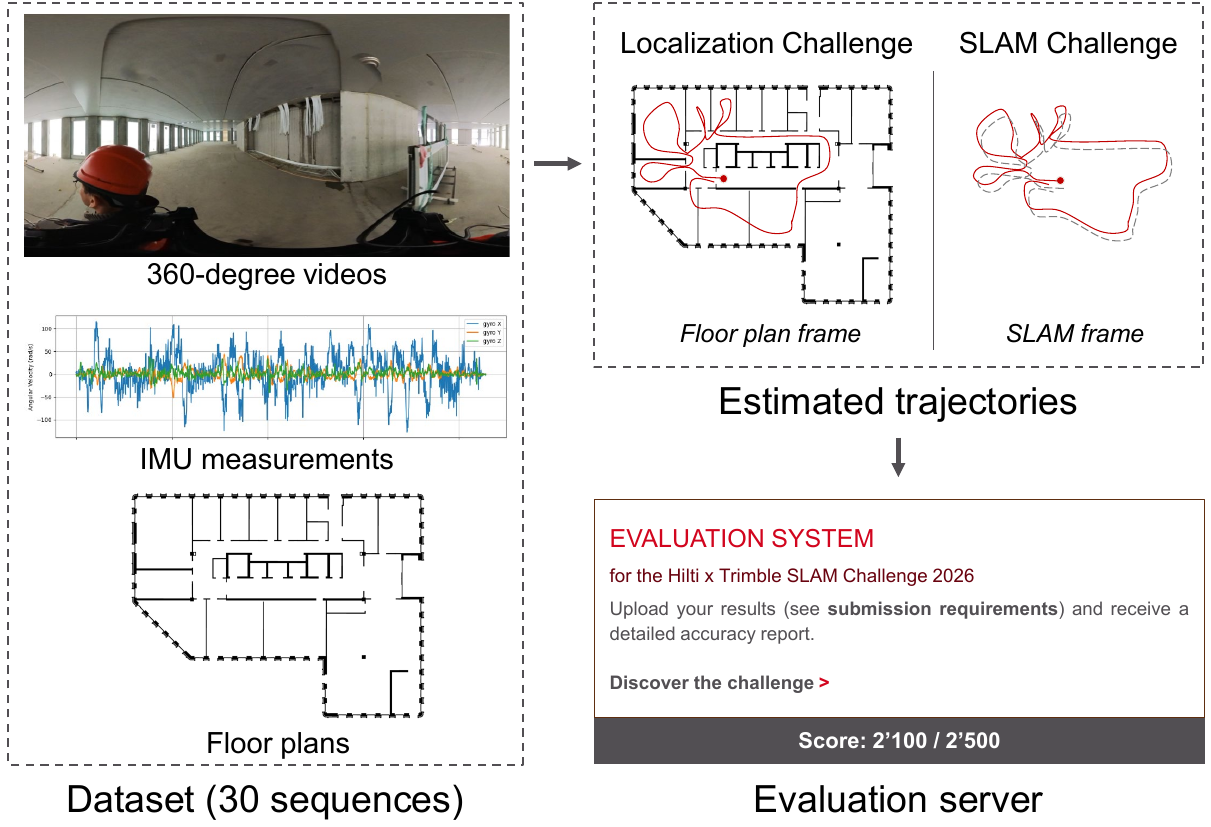}
    \caption{The Hilti-Trimble-Oxford Dataset, consisting of 360-degree videos, IMU measurements and floor plans, allows for benchmarking algorithms for SLAM and floor-plan-referenced localization.}
    \label{fig:teaser}
\end{figure}

\IEEEpubidadjcol
The core technology allowing for the generation of such maps is simultaneous localization and mapping (SLAM), which is an area of active research in robotics and computer vision~\cite{cadena2016slam}. While LiDAR sensors remain widely used for SLAM, color cameras provide a richer understanding of the environment. Due to their low cost, wide availability, and easy deployment, passive cameras are increasingly used as the sole sensor for data collection. In particular, 360 cameras provide a full field-of-view coverage, which enables efficient environment mapping. Additionally, many 360 cameras incorporate inertial measurements, which can support gravity alignment and scale estimation.

However, maps and trajectories obtained from different data-collection sessions are not sufficient on their own. For longitudinal monitoring, they need to be registered to a common reference frame, such as the construction site plan.
Prior work focuses localization within complete 3D building representations~\cite{huang2025slabim_dataset}, though building information modeling (BIM) models are not universally available, and tend to be large and complex to process. In contrast, 2D floor plans are widely adopted, compact, and still encode important geometric and semantic information about the environment. This makes floor plan-based localization a practical and relevant problem for construction-site monitoring.

This setting highlights the need for benchmark datasets that evaluate the spatial localization capabilities underpinning construction site monitoring, including visual-inertial SLAM and localization with respect to 2D floor plans in realistic, evolving construction environments.
To address this gap, we introduce the Hilti-Trimble-Oxford Dataset and the corresponding Hilti \texttimes{} Trimble SLAM Challenge 2026, which provide data, ground truth, and an evaluation protocol for benchmarking trajectory estimation methods, as shown in Fig.~\ref{fig:teaser}.

The main contributions of this work are:

\begin{itemize}
    \item The Hilti-Trimble-Oxford Dataset of $30$ challenging sequences of $360^\circ$ field of view image streams with synchronized IMU measurements, captured over a span of $8$ months at the same construction site, together with the corresponding 2D floor plans.
    \item Ground truth trajectories collected with a high-accuracy LiDAR-inertial system which are released alongside the dataset.
    \item An evaluation system for benchmarking algorithms from two categories, where participants estimate the trajectory of the 360 camera: (i) in an arbitrary reference frame (\textbf{SLAM} category) (ii) in the floor plan's reference frame (\textbf{Localization} category).
    \item The discussion of the results of the Hilti \texttimes{} Trimble SLAM Challenge 2026, which received submissions from 22 teams in the Localization category and 62 teams in the SLAM category, demonstrating substantial interest in the proposed benchmark.
\end{itemize}

\section{Related Work}
\label{sec:relatedwork}
Research into visual SLAM has progressed hand in hand with the availability of high quality datasets. Accordingly, we review datasets for visual SLAM and localization before turning to datasets for construction monitoring.

\subsection{Datasets for Visual SLAM and Localization}
Common visual SLAM datasets exist for 
autonomous vehicles~\cite{kitti}, aerial robots~\cite{Burri2016euroc}, and legged robots~\cite{wei2024fusionportablev2}.
The manner in which the sensors are mounted on the platforms and how they move has a significant impact on the usefulness of any dataset. Early autonomous vehicle and field robotics datasets, such as KITTI~\cite{kitti}, mostly contain forward translation with low rotation rates and larger distances to nearby objects. With the advent of body-mounted or egocentric sensing systems, there are new challenges for visual SLAM systems to solve, notably the high rotation rates which occur when a person moves their head abruptly.
Subsequent datasets have included inertial data and were used to develop visual-inertial odometry systems~\cite{Burri2016euroc, schubert2018tumvi}. More recent datasets such as LaMAR~\cite{sarlin2022lamar} target egocentric scenarios, while LaMAria~\cite{Krishnan_2025_ICCV} extends this approach to city-scale large environments.

For evaluation of visual SLAM systems, ground truth poses can be obtained from GNSS (especially in autonomous driving~\cite{kitti}) or from motion capture systems in indoor environments~\cite{sturm2012slambenchmark}. Control points have also been used to provide precise but sparse ground truth locations~\cite{zhang2022hilti,Krishnan_2025_ICCV}. In addition to ground truth trajectories, there are datasets that also provide a ground truth 3D map to evaluate the SLAM reconstruction. Datasets including Matterport3D~\cite{Matterport3D}, ScanNet~\cite{dai2017scannet} and LaMAR~\cite{sarlin2022lamar} use the output from an accurate SLAM system as the ground truth map. Alternatively, a survey-grade laser scanner can produce more precise 3D maps, and is used to provide ground truth maps in datasets including ETH3D~\cite{schops2017eth3d}, ScanNet++~\cite{yeshwanthliu2023scannetpp}, FusionPortableV2~\cite{wei2024fusionportablev2} and Oxford Spires~\cite{tao2025spires}.

Visual localization or place recognition (in a prior map) is a common and related problem. Early place recognition datasets such as 7-Scenes~\cite{shotton2013scene} were limited to small static scenes. Later efforts pushed the scale of viewpoint, illumination and structural change. InLoc~\cite{taira2018inloc} features textureless indoor areas while Aachen Day-Night~\cite{sattler2018benchmarking} has sequences captured with extreme variations in image appearance. However, these approaches localize in prior maps of visual features as opposed to localizing directly within an architectural floor plan, which is the focus of the Hilti-Trimble-Oxford Dataset.

\subsection{Datasets Specific to Construction Monitoring}
Despite the emergence of a wide range of technologies for digitizing construction project monitoring, there are few high quality public SLAM datasets available. This is in part because site owners and contractors prefer to retain confidentiality on their projects. The following datasets therefore provide important benchmarks, but differ in terms of sensing modalities, prior information, and the type of localization or mapping task they support.

ConSLAM is a dataset of five sequences collected with a portable LiDAR scanner on the same part of a construction site over an extended period~\cite{trzeciak_conslam}, which supports multi-session LiDAR SLAM. However, its limited spatial coverage and focus on LiDAR sensing make it less suitable for evaluating low-cost visual localization approaches.

Hilti Research has organized a series of SLAM Challenges focused on benchmarking performance in realistic construction environments. The initial Hilti SLAM Challenge (2021)~\cite{Helmberger_2022} introduced a multi-sensor dataset combining visual, LiDAR, and inertial measurements with millimeter-accurate ground truth, highlighting the importance of sensor fusion for robustness in real-world scenarios. Follow-on Challenges in 2022 and 2023 extended the benchmark to more diverse environments, sparse survey-grade ground truth, multi-device setups, and multi-session SLAM~\cite{zhang2022hilti,Nair_2024}. Across these Challenges, reported results showed strong performance from LiDAR-based methods, including sub-centimeter drift in some 2022 sequences, while also revealing that multi-session capabilities remain comparatively less mature. The associated datasets, however, lack explicit map priors, which have the potential to improve the reliability of SLAM solutions.

The SLABIM~\cite{huang2025slabim_dataset} dataset is notably relevant. The authors focus on direct LiDAR localization inside a detailed 3D BIM---in this case the HKUST campus. While industry trends are promoting the use of BIMs, we feel that a useful commercial solution should not rely on such a detailed prior map for localization. The dataset proposed in the Hilti \texttimes{} Trimble SLAM Challenge 2026 instead targets visual-only localization with respect to a minimal 2D floor plan. This combination is more practical due to the low power and cost requirements of the $360$ camera we selected.

\section{Hardware}

The Insta360 ONE RS 1-Inch Edition is selected as the primary sensor for its high-quality omnidirectional imaging and portability. Its compact, self-contained design enables rapid and reliable deployment in confined, cluttered, and unstructured construction and industrial environments.

This device captures front and rear hemispherical images, each providing a field of view (FoV) exceeding $180^\circ$.
The camera also integrates a 6-axis inertial measurement unit (IMU) that provides high-frequency motion measurements. These can be used to estimate metric motion and to carry out motion compensation of the rolling shutter images.

\subsection{Calibration}
\begin{figure}[t!]
    \centering
    \scalebox{1}[-1]{%
    \begin{tikzpicture}[
        spy scope={
            magnification=4,
            size=2.5cm,
        },
        every spy on node/.style={
            draw=orange,
            line width=1.5pt
        },
        every spy in node/.style={
            draw=orange,
            line width=1.5pt
        },
        spy connection path={
            \draw[orange, line width=1.5pt] (tikzspyonnode) -- (tikzspyinnode);
        }
    ]
        \node[anchor=south west, inner sep=0] (img) at (0,0) {
            \includegraphics[width=\columnwidth]{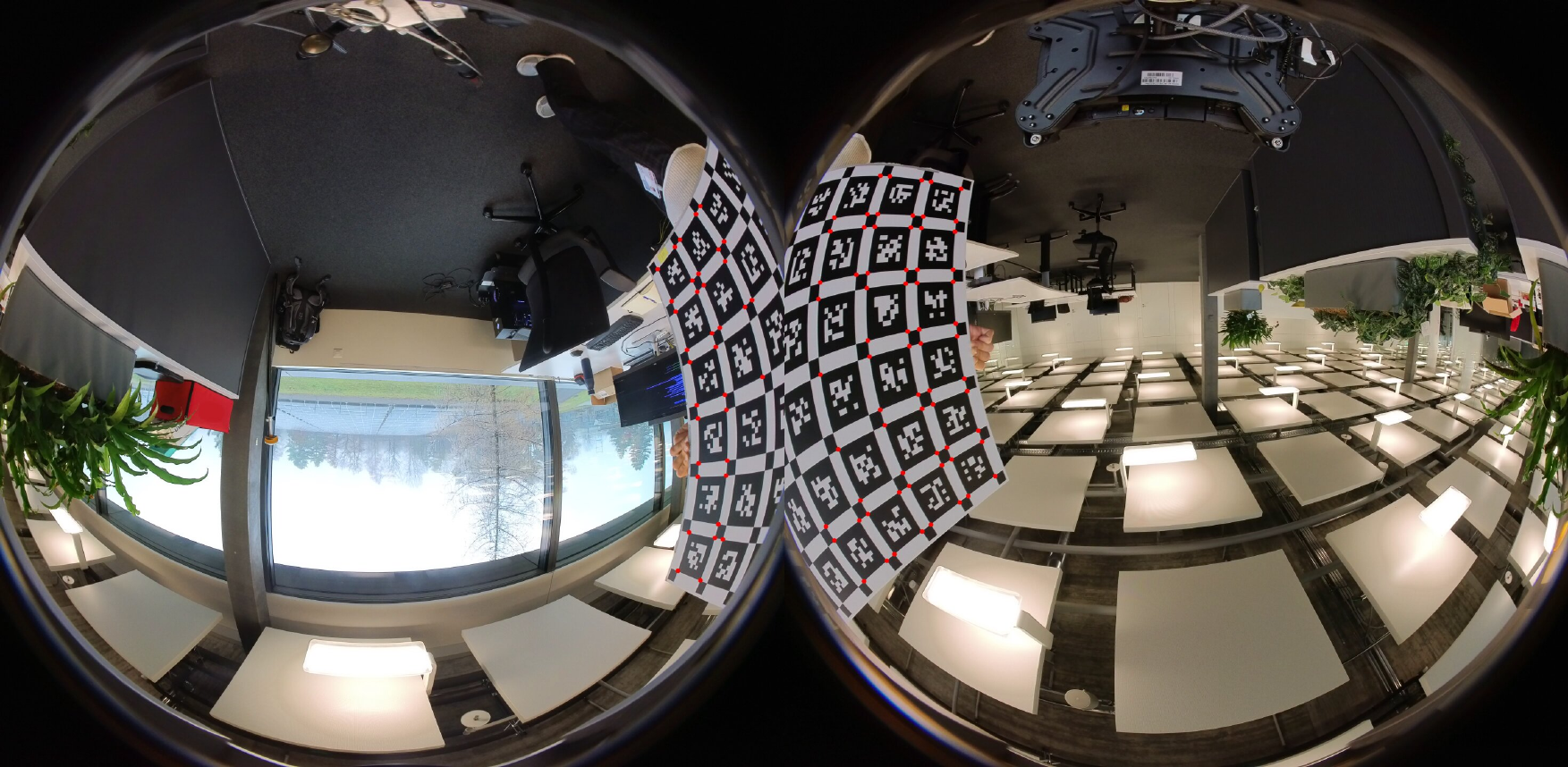}
        };

        \begin{scope}[
            x={(img.south east)},
            y={(img.north west)}
        ]
            \spy on (5.0,2.53)
                in node[anchor=south west] at (0.05,0.05);
        \end{scope}
    \end{tikzpicture}
    }

    \caption{Visualization of the multi-camera calibration. Extrinsic constraints are obtained when the AprilTag board is observed by two cameras simultaneously. The red markers indicate the forward-projected image-plane locations of the extracted calibration target corners under the calibrated cameras.}
    \label{fig:intrinsics_validation}
\end{figure}


\subsubsection{Multi-Camera Intrinsic and Extrinsic Calibration}
We employ the open-source toolbox Kalibr \cite{kalibr} for intrinsic and extrinsic calibration of two fisheye cameras. Due to the limited visual overlap between the two lenses, standard checkerboard patterns are not suitable, as they require both cameras to simultaneously observe a large, contiguous set of shared inner corners. Instead, we use a $6\times6$ AprilTag calibration board. We keep the  device stationary while moving the board around the field of view of each camera to ensure spatial coverage and feature diversity, as shown in Fig.~\ref{fig:intrinsics_validation}. We move the calibration board slowly to ensure that the rolling shutter effect is negligible. 

To achieve stable and accurate calibration under wide-angle imaging conditions, we adopt the enhanced unified camera model (EUCM)~\cite{khomutenko2015eucm}. Since many SLAM systems support the Kannala-Brandt four-parameter model (KB4)~\cite{equidistant2006distortion}, we also provide KB4 parameters fitted to the calibrated EUCM intrinsics. Fig.~\ref{fig:intrinsics_validation} shows an example of the reprojection of the extracted calibration target corners using the calibrated intrinsics and extrinsics.

\subsubsection{Rolling Shutter Parameter}
We obtain the rolling-shutter frame readout time directly from the camera metadata and validate it using a high-frequency blinking LED, where the spacing of the induced horizontal brightness bands provides an independent estimate of the sensor readout time.

\subsubsection{IMU Calibration}
We estimate the IMU noise characteristics through Allan variance analysis. Specifically, we collected approximately three hours of stationary IMU data to compute the noise density and random walk parameters for both the accelerometer and the gyroscope.

\subsubsection{Camera-IMU Extrinsic Calibration}
Given the calibrated camera and IMU noise parameters, we perform camera–IMU extrinsic calibration using Kalibr with a recording that includes rich translational motion and rotational excitation about the yaw, pitch, and roll axes to ensure full observability of all extrinsic parameters.

\section{Dataset}

The dataset spans an eight-month period at a construction site in Buchs, Switzerland. This long collection period captures the site's structural evolution, offering a novel benchmark for long-term mapping in dynamic environments. The building footprint is approximately $40\,\mathrm{m} \times 100\,\mathrm{m}$ and spans eight floors as well as two underground parking levels. In its early stages, the environment is characterized by limited texture and repetitive concrete structures; as construction progresses, the scene complexity increases as ducting, fixtures and dividing walls are installed.

We provide the dataset as ROS 2 bag files (\texttt{.db3}), a standard logging format for time-stamped sensor streams commonly used in robotics.
Raw images and high-frequency IMU measurements are extracted directly from the \texttt{.insv} file (Insta360 native format for storing recorded video and associated metadata). Each recording is self-contained and captured on a single floor, enabling modular benchmarking and evaluation.
Specifically, each ROS 2 bag records a set of time synchronized data streams (``topic'') as summarized in Table~\ref{tab:ros_topics}. These topics provide the complete sensor data required for visual-inertial SLAM, including IMU measurements and two fisheye camera streams.

\begin{table}[t!]
    \centering
    \caption{Sensor ROS Topics and Frequencies.}
    \begin{tabular}{ l | l | c}
        Sensor & Topic name & Rate [Hz] \\ 
        \hline
        IMU & \texttt{/imu/data\_raw/compressed} & 1000 \\ 
        Front camera & \texttt{/cam0/image\_raw/compressed} & 30 \\ 
        Rear camera & \texttt{/cam1/image\_raw/compressed} & 30    
    \end{tabular}
    \label{tab:ros_topics}
\end{table}

\subsection{Challenge Sequences}
\label{sec:challenge_sequences}
The dataset consists of $30$ sequences with an average duration of \SI{180}{\second}. We categorize the runs according to five factors that influence sequence difficulty:

\begin{enumerate}
    \item \textbf{Clutter:} Selected runs offer intentionally cluttered environments (e.g. close proximity to walls or moving people), resulting in severe occlusions.
    \item \textbf{Low-Light:} Sequences include near-dark conditions, testing the limits of feature tracking.
    \item \textbf{Initialization:} \textit{Static} runs start with a stationary period and gentle excitation for IMU bias estimation. \textit{Dynamic} runs begin in motion, requiring the estimator to initialize under active dynamics.
    \item \textbf{Aggressive Motion:} Certain runs involve rapid $6$-DoF movements, introducing motion blur and stress-testing IMU-based de-skewing. These sequences may require rolling shutter compensation for optimal results.
    \item \textbf{Poor texture:} Captures contain regions with weak visual structure, reducing the quality of feature tracking.
\end{enumerate}

By combining these factors, the dataset spans diverse operating conditions, as summarized in Table~\ref{tab:sequence_conditions}, ranging from simple cases with static initialization, sufficient illumination, rich visual features, and smooth motion, to challenging cases with dynamic initialization, low-light or near-dark scenes, aggressive 6-DoF motion, and severe visual occlusions.

The sequences intentionally avoid large repeated loops, making loop-closure detection and global map optimization more difficult and emphasizing wide-field-of-view loop closure using both cameras.

\begin{table}[t!]
    \centering 
    \caption{Characterisation of challenge sequences.}
    \resizebox{\columnwidth}{!}{%
        \begin{tabular}{r|cccccccccccc}
            \toprule
            \textbf{Clutter}
            & & & & & & & & & & \checkmark & & \checkmark \\
        
            \textbf{Low light}
            & & & & & \checkmark & \checkmark & & \checkmark & \checkmark & \checkmark & \checkmark & \checkmark \\

            \textbf{Dynamic init.}
            & & & & \checkmark & & \checkmark & \checkmark & & & & \checkmark & \checkmark \\
        
            \textbf{Aggr. motion}
            & & & \checkmark & & & & & \checkmark & & \checkmark & \checkmark & \checkmark \\
        
            \textbf{Poor texture}
            & & \checkmark & & & & & \checkmark & & \checkmark & & & \\
        
            \midrule
            \textbf{\# Sequences}
            & \textbf{6}
            & \textbf{4}
            & \textbf{2}
            & \textbf{4}
            & \textbf{1}
            & \textbf{2}
            & \textbf{3}
            & \textbf{1}
            & \textbf{3}
            & \textbf{2}
            & \textbf{1}
            & \textbf{1} \\
            \bottomrule
        \end{tabular}%
    }

    \vspace{3pt}
    \parbox{\columnwidth}{\footnotesize
        \textit{Note:} Each column represents a unique combination of conditions, with the number of sequences shown in the bottom row. Sequence properties and videos
        are available on the dataset webpage.
    }
    \label{tab:sequence_conditions}
\end{table}

\subsection{Floor Plans} \label{sec:floor_plan}
\begin{figure}
    \centering
    \includegraphics[width=\linewidth]{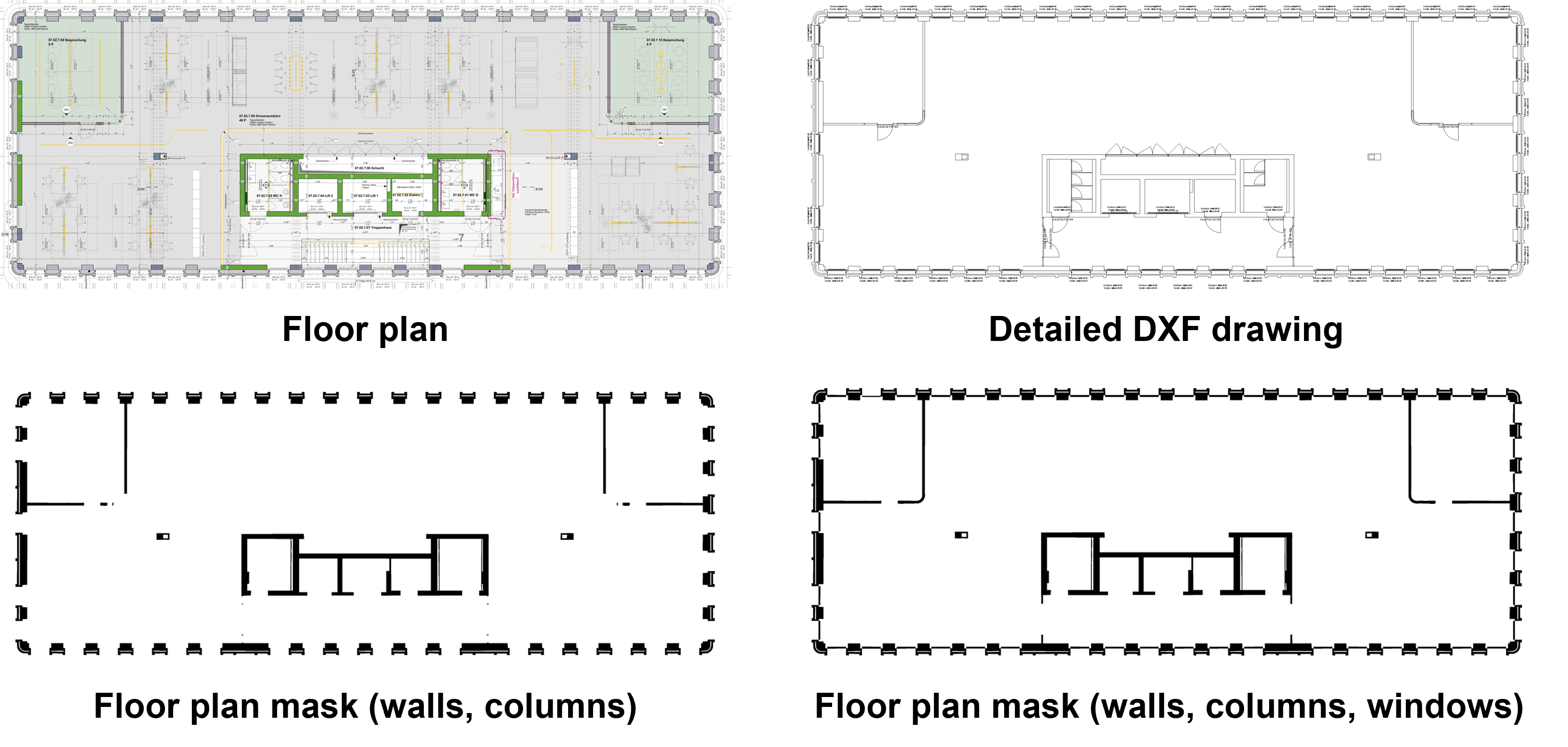}
    \caption{Types of provided floor plans.}
    \label{fig:floorplans}
\end{figure}

We provide anonymized building floor plans as PNG images, resized to a metric resolution of $1\,\mathrm{px}$ = $1\,\mathrm{cm}$ and simplified to contain only structural elements such as walls and concrete columns. The resulting maps are represented as binary images, where structural elements are encoded as 1 (black) and free space as 0 (white), allowing direct conversion into ROS occupancy maps. Two variants of each floor plan are available: with and without windows included as structural elements. These elements were manually labelled on professional architectural floor plans at a higher resolution, with an uncertainty of 1--3 pixels after scaling. Additionally, we provide floor plans in DXF format, which contain more detailed structural information. Although these are not registered with respect to the PNG maps, they offer additional geometric or semantic data. All floor plans' formats available in the dataset are presented in Fig.~\ref{fig:floorplans}.

Localization in this setting remains challenging, as it requires associating visual observations from the sequences with the simplified 2D structural representation. To reduce ambiguity and shift the focus from global place recognition to local map matching, the initial pose of each sequence within the floor plan is provided.

\section{Ground Truth}
\label{ground_truth}

\begin{figure}[t]
        \centering
        \includegraphics[width=0.6\linewidth]{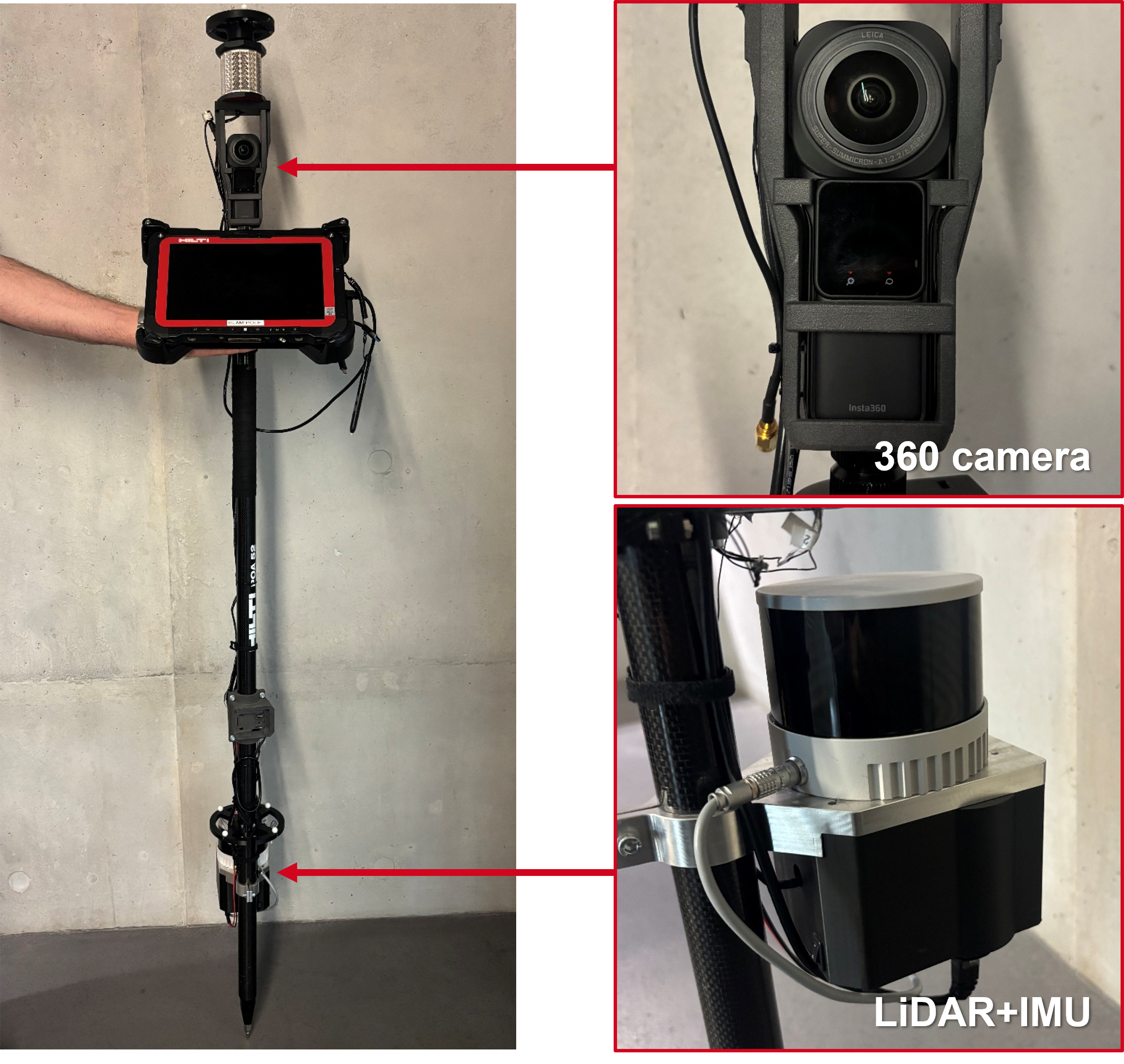}
        \caption{Data collection platform: rigidly connected V\&R LiDAR-inertial SLAM system and Insta360 ONE RS 1-Inch 360 Edition.}
        \label{fig:slampole}
    \end{figure}

    We obtain ground truth for both camera trajectories and the reconstructed scene geometry using a LiDAR-inertial SLAM system from our industrial partner, rigidly attached to the 360 camera as depicted in Fig.~\ref{fig:slampole}. The LiDAR-inertial subpart consists of a Hesai XT32M2X LiDAR sensor and a high-quality, factory-calibrated, tactical-grade IMU, with hardware synchronization between the two components.
    
\subsection{Temporal Synchronization} \label{sec:temporal_synchronization}
    While the Insta360 and the LiDAR-inertial system are each internally hardware-synchronized with their own IMU, they are not synchronized with each other. We temporally align the two IMUs by correlating gyroscope magnitudes to initialize the time offset, then refine a linear timestamp mapping $t_L= \alpha t_C + \Delta t$, where $\alpha$ is the clock-scale factor, $\Delta t$ is the time offset and $t_L, t_C$ correspond to LiDAR and camera timestamps respectively. The resulting time mapping is used to convert timestamps between the two clock domains.

\subsection{LiDAR--Inertial Reference Trajectory} \label{sec:reference_trajectory}
    Ideally, the reference trajectories would be independently measured using survey-grade terrestrial laser scanners or surveyed control points. However, such measurements were not possible due to the pace of the ongoing construction project. We therefore use trajectories estimated by a high-quality LiDAR--inertial SLAM system as the reference. The underlying technology has demonstrated centimeter-level accuracy (${\approx}3.94\,\mathrm{cm}$), and sub-centimeter accuracy ($0.7\,\mathrm{cm}$) under favorable conditions, in previous construction-site SLAM Challenges with independently surveyed ground truth~\cite{zhang2022hilti}.
    Compared to the consumer-grade IMU in the Insta360 camera, the reference system combines a Hesai LiDAR with a factory-calibrated tactical-grade IMU and performs full offline optimization over all recorded measurements using a continuous-time trajectory representation. Given that the errors of the best visual--inertial submissions in the present benchmark are substantially larger than the expected centimeter-scale error of this reference system, we consider the resulting trajectories sufficiently accurate for the intended evaluation.

\subsection{Camera Ground Truth} \label{sec:camera_ground_truth}
    To transfer the LiDAR trajectory to the camera frame, we perform the LiDAR--camera extrinsic calibration. Given the calibrated camera intrinsics, an AprilTag grid is moved to multiple poses observed by both sensors. From the camera detections and the known physical dimensions of the printed target, we compute the metric pose of the calibration plane for each placement. Starting from a manually defined initial LiDAR--camera transform, the optimization incorporates two complementary error terms for joint optimization: (i) the reprojection error between detected AprilTag corners and their predicted image locations, and (ii) point-to-plane distances between LiDAR measurements and the observed calibration board, which is assumed to be planar. While we use an internal calibration tool from our industrial partner, we note that a similar open-source tool is available~\cite{zhi2022multical}.

\subsection{Floor Plan Alignment}

\begin{figure}[t]
        \centering
        \includegraphics[width=0.7\linewidth]{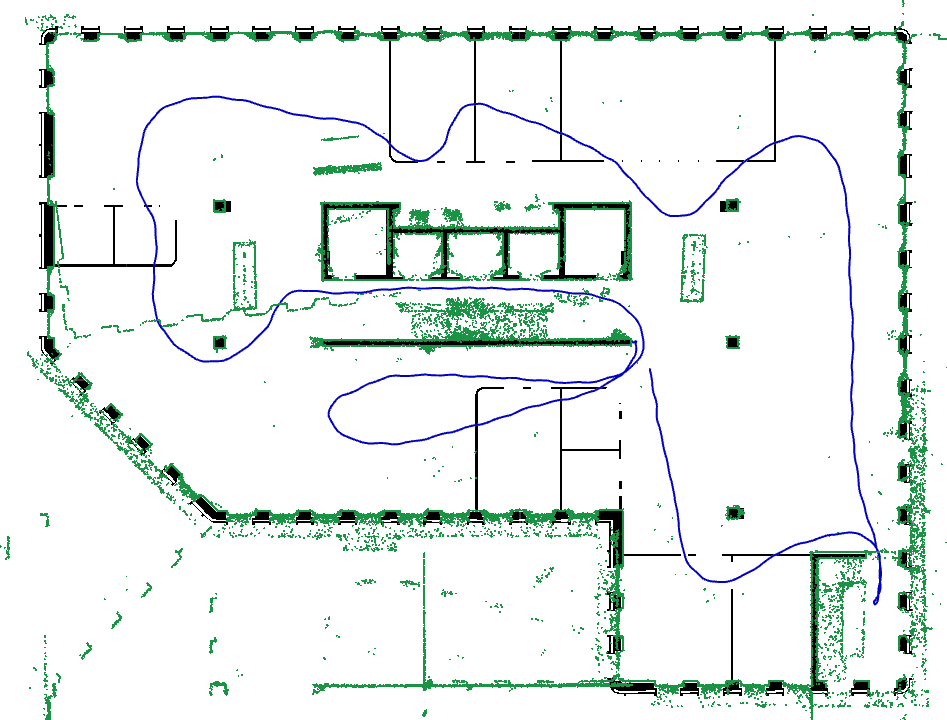}
        \caption{A ground truth trajectory (\textcolor{darkblue}{\textbf{blue}}) with LiDAR SLAM map (\textcolor{darkgreen}{\textbf{green}}) overlaid on a floor plan (\textbf{black}). The trajectory passes through some as-yet unbuilt walls due to the early construction phase, and some temporary wall-like structures are captured in the LiDAR point cloud. Glass reflections can also be seen.} 
        \label{fig:slam_system}
    \end{figure}

    We achieve the spatial alignment of the trajectories to the architectural floor plans through a multi-stage registration process. Initially, we perform a coarse alignment by manually rotating and translating the LiDAR point cloud to match the primary structural features of the floor plan. This result serves as the initial seed for a fine-grained refinement using the point-to-point iterative closest point (ICP) algorithm \cite{besl1992method}, ensuring a high-precision fit between the global LiDAR map and the reference geometry (Fig.~\ref{fig:slam_system}).
    
    Beyond alignment, this process quantifies the discrepancies between the as-built state and the as-planned architectural models. We observe an average deviation of approximately $2\,\mathrm{cm}$ between the LiDAR data and the floor plans. These values reflect the standard structural tolerances and inevitable variances inherent in large-scale, real-world construction environments. Thus, differences of a few centimeters between methods should not be overinterpreted.
    
\section{Challenge Results and Findings}

\subsection{Benchmark Overview}
\label{sec:results:overview}

    The proposed dataset is accompanied by a benchmarking system that evaluates submitted trajectories according to the selected evaluation track:

    \smallskip\noindent
    \textbf{SLAM:} This track follows standard SLAM evaluation. The submitted and ground truth trajectories are first rigidly aligned in a least-squares sense using the Kabsch algorithm~\cite{Kabsch1976}. The score is then computed from the resulting 3D absolute trajectory error (ATE).
    
    \smallskip\noindent
    \textbf{Localization:} This track requires localization with respect to the floor plan. The submitted trajectory has to be expressed in the map frame (without any alignment). In this track, only the 2D ATE in the floor plan plane is evaluated, with the $z$-coordinate ignored.
    \smallskip
    
    To encourage submissions to the benchmark, we organized the Hilti \texttimes{} Trimble SLAM Challenge 2026, which accepted submissions over a period of 3.5 months. After the completion of the Challenge, all ground truth trajectories were made available to support independent evaluation, further algorithm development, and model training.

    \begin{table*}[t!]
        \centering
        \caption{Comparison of the top five SLAM teams.}
        \begin{tabular}{clllccccccccc}
            \toprule
            & Team & Base method & View type & \makecell{Map\\prior} &   \makecell{Real\\time} & \makecell{Global\\BA} & \makecell{Loop\\closure} & \makecell{Same\\params} & \makecell{Average\\RMSE [m]} & Score \\
            \midrule
            1 & KAIST & OKVIS2-X & Fisheye & \xmark & - & \checkmark  & \checkmark & - & 0.089 & 2410.0 \\
            2 & Inha Univ.& $\sqrt{\mathrm{VINS}}$ & Fisheye & \xmark & \xmark & \checkmark & \checkmark & - & 0.109 & 2393.9 \\
            3 & - & OKVIS2-X & Fisheye & \xmark &  \xmark & \checkmark & \checkmark & \xmark & 0.189 & 2320.7 \\
            4 & ICL & OpenVINS + DBoW & Fisheye & \xmark &  \xmark & \xmark & \checkmark & \checkmark & 0.202 & 2308.8 \\
            5 & Univ. of Luxembourg & ORB-SLAM3 & Fisheye & \checkmark &  - & \checkmark & \checkmark & - & 0.224 & 2299.3 \\
            \bottomrule
        \end{tabular}
        
        \vspace{3pt}
        \parbox{\linewidth}{\footnotesize
            \textit{Note:} Teams are ranked by their challenge results. The comparison includes the base SLAM method, image view type, use of map priors, real-time capability, global bundle adjustment, loop closure, parameter consistency across sequences, average RMSE, and overall score.
        }
        \label{tab:results_slam}
    \end{table*}

    \begin{table*}[t!]
        \centering
        \caption{Comparison of the top five localization teams.}
        \begin{tabular}{llllccc}
            \toprule
            &Team & Trajectory estimation approach & Map alignment approach & \makecell{Semantic \\segment.} & \makecell{Average\\RMSE [m]} & Score \\
            \midrule
            1 & ETH-THU-Stuttgart & OKVIS2-X & Z-FLoc anchors & \checkmark~\cite{cheng2021mask2former} & 0.238 & 2196.3 \\
            2 & Univ. of Luxembourg & ORB-SLAM3 with map priors & BEV-wall-points-to-map ICP in 2D & \checkmark~\cite{kerssies2025eomt} & 0.286 & 2161.4\\
            3 & ETH OmniRecon & OpenVINS + COLMAP & BEV-wall-points-to-map ICP in 2D & \checkmark~\cite{cheng2021mask2former} & 0.399 & 2041.3\\
            4 & Cairo University & ORB-SLAM3 & Ray-based wall scoring & \xmark & 0.497 & 1919.0\\
            5 & ETH Map-It Ralph & OpenVINS + COLMAP & Starting-pose-based alignment & \xmark & 1.287 & 1569.9\\
            \bottomrule
        \end{tabular}

        \vspace{3pt}
        \parbox{\linewidth}{\footnotesize
            \textit{Note:} Teams are ranked by their challenge results. The comparison includes the trajectory generation method, trajectory-to-map alignment approach, use of video-based semantic segmentation for alignment, average RMSE, and overall score.
        }
        \label{tab:results_localization}
    \end{table*}

\subsection{Evaluation Metrics}
\label{sec:metric}
    
    We required each submission to provide a trajectory with one pose for each camera-frame timestamp. A run is evaluated only if the submitted trajectory achieves at least $99\,\%$ coverage of the evaluated ground truth poses, with coverage defined as the ratio of matched predicted poses to ground truth poses; otherwise, the score for that run is set to zero.  The initial \SI{5}{\second} of each recording are excluded to ignore any effects during initialization. 
    
    For each evaluated trajectory $i$, we compute an exponentially decaying position score by averaging the per-pose scores over the trajectory:
    \begin{align}
        \mathrm{score}_i
        &= \frac{1}{N_i}\sum_{t=1}^{N_i} a e^{-c\, e^{(i)}_t} \label{eq:score_i},\\
        e^{(i)}_t
        &= \left\lVert \mathbf{p}^{(i)}_{\mathcal{M},\mathcal{C}}(t) - \hat{\mathbf{p}}^{(i)}_{\mathcal{M},\mathcal{C}}(t) \right\rVert_2.
    \end{align}
    Here, $N_i$ denotes the number of evaluated poses in trajectory $i$; $\mathbf{p}^{(i)}_{\mathcal{M},\mathcal{C}}(t)$
    and $\hat{\mathbf{p}}^{(i)}_{\mathcal{M},\mathcal{C}}(t)$ denote the ground truth and predicted camera position in the map frame $\mathcal{M}$; and $e^{(i)}_t$ is the Euclidean position error. For the SLAM track, $e^{(i)}_t$ is the 3D Euclidean distance after rigid alignment. For the Localization track, $e^{(i)}_t$ is the 2D Euclidean distance in the floorplan plane.
    The constants $a$ and $c$ parameterize the exponential decay and are chosen such that an error of \SI{0}{\meter} corresponds to a score of $100$ and an error of \SI{10}{\meter} corresponds to a score of $1$, yielding $a=100$ and $c=\ln(10)/5 \approx 0.4605$.
    The final score is the sum of the trajectory scores over all runs:
    \begin{align}
        \mathrm{finalScore} = \sum_i \mathrm{score}_i.
    \end{align}
    
    For the official benchmark, both the SLAM and Localization tracks exclude $5$ sequences that were made available with ground truth. These sequences are intended solely for algorithm fine-tuning and not used for evaluation. Therefore, the SLAM track is evaluated on $25$ trajectories out of $30$ in total. The Localization track is evaluated on $24$ trajectories, excluding the same $5$ runs as SLAM, with one additional sequence omitted due to a missing floor plan. Consequently, the maximum achievable total score is $2500$ points for SLAM and $2400$ points for Localization. All the aforementioned metrics are computed using the \texttt{evo} toolbox~\cite{grupp2017evo}.

\subsection{Results and Leading Approaches} \label{sec:results:results}
    At the time of the Challenge deadline, the evaluation system has received over $1800$ submissions, with 22 unique teams competing in the Localization track and 62 teams in the SLAM track. The results of the Hilti \texttimes{} Trimble SLAM Challenge 2026 were announced publicly and the top performing solutions are shown in Table~\ref{tab:results_slam} and Table~\ref{tab:results_localization}, with a dash signifying the lack of information provided by the participants. The full leaderboard and detailed technical reports from each team are available on the Challenge webpage.

    The winning SLAM submission, ACDC-VSLAM from KAIST, achieved a total score of $2410.0$ points. Their system is based on OKVIS2-X~\cite{boche2025okvis2x} and introduces several adaptations tailored to the benchmark. In the frontend, spherical equal-area feature selection improves angular coverage of the fisheye images, while line features provide additional constraints in texture-poor regions. Its local bundle adjustment further re-weights visual constraints based on track reliability and depth. In the backend, a two-stage loop-closure module detects both intra-camera and cross-camera revisits, and adaptively selects between $\mathrm{SE}(3)$, $\mathrm{Sim}(3)$, and 4-DoF corrections according to the quality of each loop candidate. These design choices directly target the wide-FoV distortion, weak texture, and front--rear camera configuration present in the dataset.
    Other high-performing SLAM submissions use complementary strategies to address similar challenges, including low-light enhancement and dynamic-object masking, point-and-line feature fusion, cross-camera constraints, robust initialization, and global bundle adjustment or factor-graph optimization.

    A joint team from ETH, THU, and University of Stuttgart won the Localization track with the score of $2196.3$ points. Their approach first estimates camera poses using OKVIS2-X with online extrinsic calibration. They then build upon Z-FLoc~\cite{umemura2026zfloc}, which estimates a $\mathrm{Sim}(2)$ transformation from recent image frames. It first uses semantic segmentation to extract structurally stable wall and floor regions, reconstructs a wall-only 3D point cloud, and projects it into a bird's-eye-view (BEV) map. It then matches extracted line/circular primitives to floor plan primitives with RANSAC and non-linear refinement, and refines the overall trajectory using pose graph optimization. Their solution achieves an average RMSE of $24\,\mathrm{cm}$.
    Other strong Localization submissions similarly exploit semantically stable wall structures, typically using semantic segmentation followed by BEV-based geometric alignment to the floor plan.

\subsection{Discussion} \label{sec:results:discussion}
    The top SLAM solutions all rely on visual-inertial odometry, usually using both cameras and the IMU, and followed by a subsequent optimization step (BA). Most are based on established systems such as OKVIS2-X~\cite{boche2025okvis2x}, OpenVINS~\cite{Geneva2020ICRA}, ORB-SLAM3~\cite{campos2021orbslam3}, or $\sqrt{\mathrm{VINS}}$~\cite{Peng2025sqrtvins}. Notably, OKVIS2-X was used by several high-performing teams, including the winners of both the SLAM and Localization tracks. 
    
    Some teams explicitly reported challenges with initialization, especially for sequences that start in motion, and addressed this through dynamic or delayed initialization strategies. Image preprocessing was also common, including contrast enhancement in dark scenes, static masks, and segmentation to remove the operator or dynamic objects. Beyond classical point features, multiple systems incorporate line features, which are particularly useful in construction environments with texture-poor walls, corridors, and strong structural edges. In the backend, loop closure, pose-graph optimization, and global bundle adjustment are frequently used to reduce drift beyond what local visual-inertial odometry can recover. Particularly, cross-camera loop closure is commonly adopted to improve robustness by exploiting the complementary front and rear cameras, which together provide $360^\circ$ visual coverage. These results suggest that strong performance is generally associated with combining robust visual-inertial estimation with dataset-specific frontend processing and global trajectory refinement.

    To localize within the floorplan, top-performing teams typically detect walls in the images, construct wall-only point clouds, and align their BEV map with floor plans using methods such as Z-FLoc or ICP. This suggests that performance gains can come not only from geometric alignment, but also from selecting stable, semantically meaningful structural elements.
    However, no team reported improving alignment by detecting windows or other structural components other than walls that are present in detailed drawings.
    
\begin{figure}[t]
        \centering
    
        \includegraphics[width=\columnwidth]{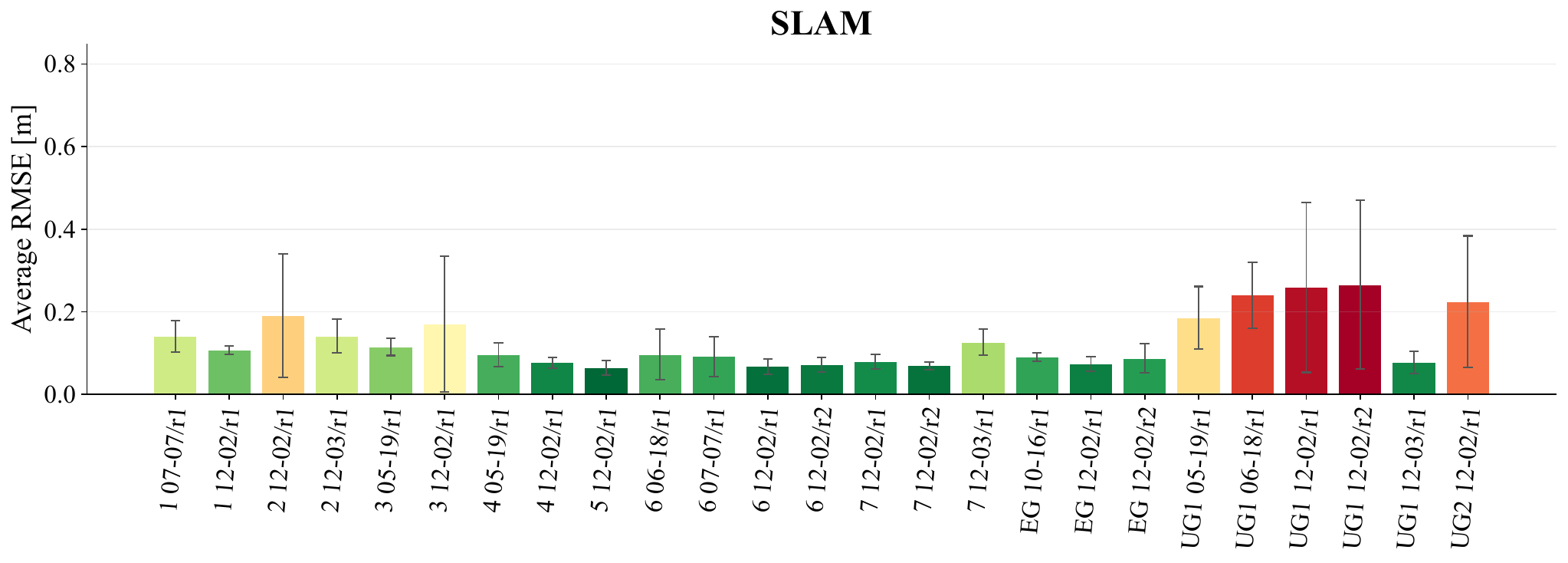}
    
        \vspace{0.5em}
    
        \includegraphics[width=0.98\columnwidth]{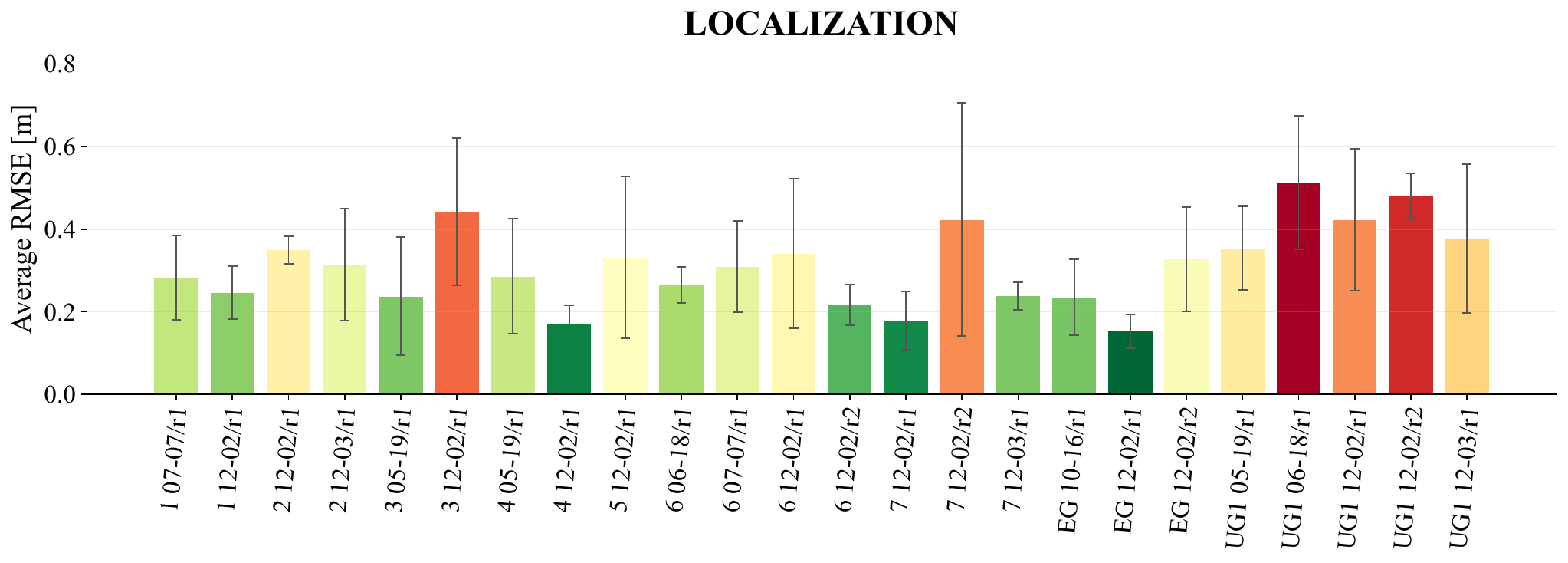}
    
        \caption{Average RMSE with standard deviations of the top three teams for each sequence in the SLAM and Localization categories. Bar color indicates error magnitude, transitioning from green to red as error increases. The coding for run names is floor, date, and numbered run on that date.}
        \label{fig:per_sequence_results}
    \end{figure}

    Sequences recorded in the underground floor consistently challenged the top teams in both the localization and SLAM tracks, as shown in Fig.~\ref{fig:per_sequence_results}. We attribute this to the combination of low-light and texture-poor conditions, larger open spaces, and longer trajectory lengths.

\subsection{Known Issues}

    There are a couple of practical considerations when using this dataset. The most important issue is the accuracy or concurrency of the provided floor plans. The bitmap floor plans were derived from a single version of the architectural floor plans. The building project was in constant progress with certain elements of the floor plan (cubicles, partitions or ducting) not yet present during the early data recordings. This means that the Challenge required localization to a similar but different scene---with only some of the features usable for matching. We feel this is a realistic challenge because up-to-the-minute floor plans would not be feasible in active construction environments.

    Separately from this, a few deviations or errors in the floor plan were identified when comparing them to our LiDAR-based ground truth point clouds (e.g., partitions installed with a slightly different length). This may lead to small residual errors in our ground truth pipeline. These effects are minor with respect to the intended evaluation scale and do not affect the overall validity of the benchmark.

\section{Conclusion}

    The Hilti-Trimble-Oxford Dataset provides a construction-focused visual-inertial state estimation benchmark with prior 2D maps. Using the Hilti \texttimes{} Trimble SLAM Challenge 2026, we evaluated SLAM and localization submissions from teams worldwide. The top team in the Localization category achieved an average error below $24\,\mathrm{cm}$, with increased performance observed among solutions that incorporate semantic scene information from video frames. The SLAM category attracted nearly three times as many teams, demonstrating high interest, with the lowest reported average error reaching $9\,\mathrm{cm}$.

    Beyond the original Challenge sequences, we have also released the ground truth trajectories and evaluation code. We hope that the Hilti-Trimble-Oxford Dataset and benchmark will support further progress in SLAM and localization research and contribute to more effective construction-site monitoring.

\section{Acknowledgements}

This paper was part funded by the UKRI/EPSRC project Mobile Robotic Inspector (EP/Z531212/1). For the purpose of open access, the author(s) has applied a Creative Commons Attribution (CC BY) license to any Accepted Manuscript version arising.

\small
\bibliographystyle{IEEEtran} 
\bibliography{string-short,references}

\end{document}